\documentclass{article}

\usepackage{arxiv}

\usepackage[utf8]{inputenc} 
\usepackage[T1]{fontenc}    
\usepackage{hyperref}       
\usepackage{url}            
\usepackage{booktabs}       
\usepackage{amsfonts}       
\usepackage{nicefrac}       
\usepackage{microtype}      
\usepackage{lipsum}
\usepackage{colortbl}
\usepackage{threeparttable}
\usepackage{graphicx}

\title{SIGVerse: A cloud-based VR platform for research on social and embodied human-robot interaction}

\author{
  Tetsunari Inamura\thanks{Use footnote for providing further
    information about author (webpage, alternative
    address)---\emph{not} for acknowledging funding agencies.} \\
  National Institute of Informatics\\
  The Gradutate University for Advanced Studies, SOKENDAI\\
  Tokyo, Japan \\
  \texttt{inamura@nii.ac.jp} \\
   \And
 Yoshiaki Mizuchi \\
  National Institute of Informatics\\
  Tokyo, Japan \\
  \texttt{mizuchi@nii.ac.jp} \\
}

\begin{document}

\maketitle

\begin{abstract}
Common sense and social interaction related to daily-life environments are considerably important for autonomous robots, which support human activities.
One of the practical approaches for acquiring such social interaction skills and semantic information as common sense in human activity is the application of recent machine learning techniques.
Although recent machine learning techniques have been successful in realizing automatic manipulation and driving tasks, it is difficult to use these techniques in applications that require human-robot interaction experience.
Humans have to perform several times over a long term to show embodied and social interaction behaviors to robots or learning systems.
To address this problem, we propose a cloud-based immersive virtual reality (VR) platform which enables virtual human-robot interaction to collect the social and embodied knowledge of human activities in a variety of situations.
To realize the flexible and reusable system, we develop a real-time bridging mechanism between ROS and Unity, which is one of the standard platforms for developing VR applications.
We apply the proposed system to a robot competition field named RoboCup@Home to confirm the feasibility of the system in a realistic human-robot interaction scenario.
Through demonstration experiments at the competition, we show the usefulness and potential of the system for the development and evaluation of social intelligence through human-robot interaction.
The proposed VR platform enables robot systems to collect social experiences with several users in a short time.
The platform also contributes in providing a dataset of social behaviors, which would be a key aspect for intelligent service robots to acquire social interaction skills based on machine learning techniques.
\end{abstract}


\section{Introduction}
Various physical functions of intelligent robots, such as learning strategies for grasping objects\,\cite{Levine2018}, manipulating flexible objects such as cloth\,\cite{Yang2017}, and automatic driving\,\cite{Grigorescu2019}, have been significantly improved by the recent development of machine learning technology.
The common assumption of these robot learning techniques is that they prepare a sample of body movements and
decisions in the form of data sets.
Some methods, such as Yang's system\,\cite{Yang2017}, require only a few tens of data.
However, the more complex is the problem, the larger is the number of data required, and the number of data is in the order of thousands to tens of thousands.
This quantity of data is not practical in real environments.
If we can create an environment in which robots can repeat the object manipulation actions, autonomous learning is possible.
However, if the target of the learning changes to the interaction behavior with humans, the situation differs.

Research on Human-Robot Interaction (HRI) is one of the most active research topics in robotics and intelligent systems.
As the HRI system is complex, there are several difficulties in the research activity.
One difficulty is the collection of a dataset for machine learning in HRI\,\cite{Amershi2014} toward learning and modeling human activity.
A typical strategy to measure the human activity in the interaction with robots is the use of a video severance system, including motion capture systems.
For example, Kanda\,\cite{Kanda2010} had developed a massive sensor network system for observing the human activity in a shopping mall environment.
The observations were performed for 25 days.
Another application of the interaction between a robot and children in an elementary school required approximately two months to collect the interaction dataset\,\cite{Kanda2007}.
The significant cost for such an observation is one of the bottlenecks in HRI research.

One of the simplest strategies for reducing data collection costs is the use of simulator systems.
Although the gap between a real environment and a simulated environment is often discussed as a common concern, a keyword named sim2real has been recently improved to close the gap.
A simulation of HRI also seems to be an attractive solution for accelerating the learning; however, another problem is that the real and natural behavior of test subjects cannot be modeled as a simulated agent in a simulation environment.

\begin{figure}[h!]
  \begin{center}
  \includegraphics[width=15cm]{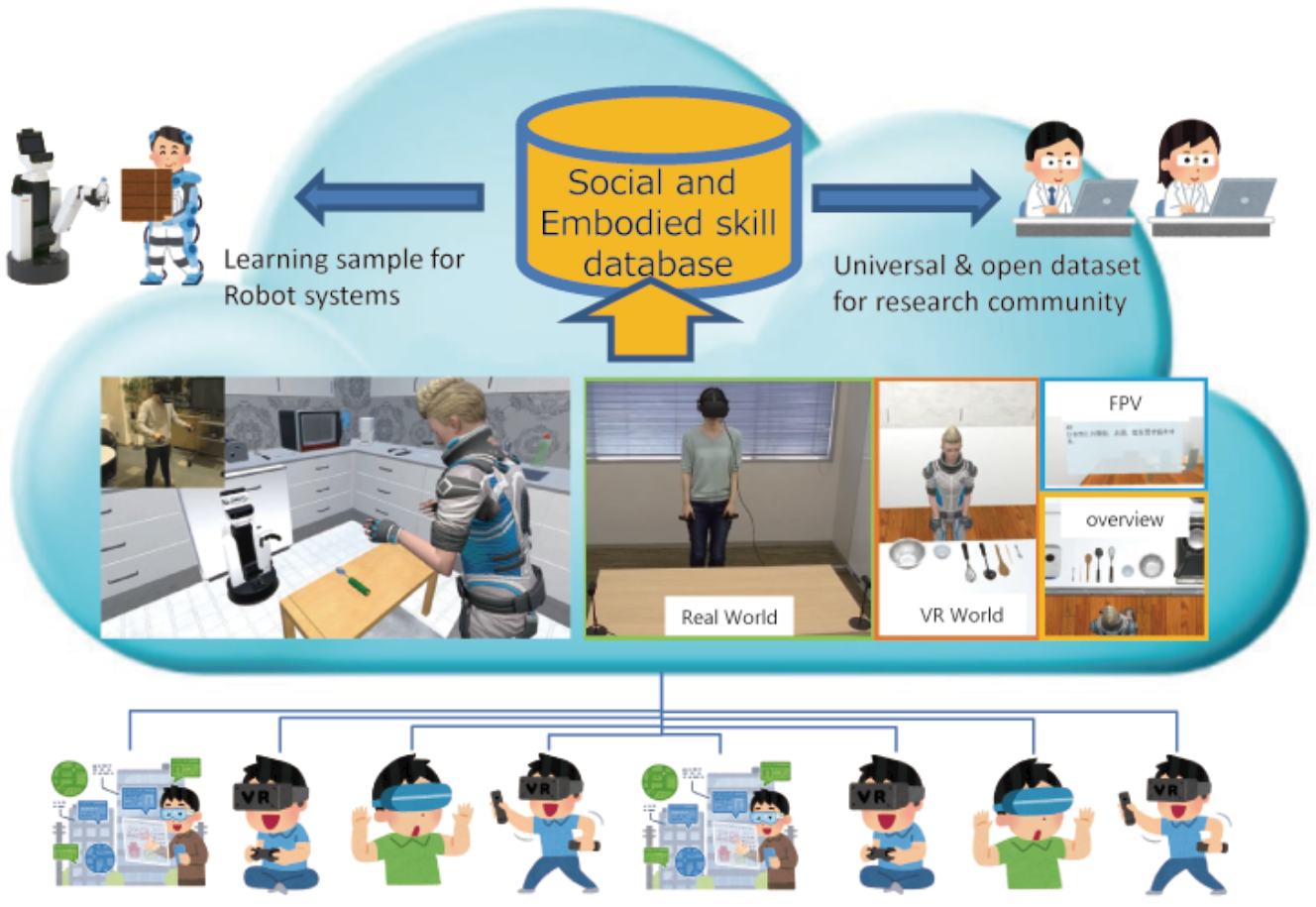}
  \end{center}
  \caption{Concept of the SIGVerse system. Arbitrary client users can participate in the HRI experiment via the Internet using VR devices.}\label{fig:1}
\end{figure}

To address these problems, we propose a concept of a cloud-based VR system for HRI shown in Fig.\ref{fig:1}.
The system allows test subjects to participate in HRI experiments, which is set up in the VR environment, via the Internet, without actually inviting them to the real laboratory or experimental field.
If the test subject can log in to an avatar with the VR device and have face-to-face interaction with the virtual robot from the first-person’s view, most of the HRI experiments that have been conducted so far can be realized in the VR environment.
The time burden on the subjects can also be distributed through crowd-sourcing, based on the invitation of more test subjects.
The purpose of this paper is to summarize the technical infrastructure required for the realization of such a system and present a system configuration of the prototype platform named SIGVerse.
We also present some case studies that show the effectiveness and feasibility of the platform and discuss the future of HRI in the VR environment.

\section{Related Works}

Bazzano et al. have developed a virtual reality (VR) system for service robots\,\cite{Bazzano2016}.
This system provides a virtual office environment and immersive VR interface to test the interaction between virtual robots and real humans.
Since the software of a service robot should be developed in C\# script on the Unity system, compatibility among real robots is low.

Lemaignan et al. have proposed SPARK, which enables social robots to assess the situation of human-robot interaction and plan proper behavior based on a spatio-temporal representation with three-dimensional (3D) geometric reconstruction from robot sensors\,\cite{Lemaignan2017}.
They also expanded the system to share the representation among a group of service robots as UNDERWORLDS\,\cite{Lemaignan2018}.
Since the main contribution of both systems is to establish the internal representation of the world for social robots, the systems do not support a user interface to share the representation with humans.
Simulation of the social interaction behavior between robot systems and real humans is an important function of service robots;
however, the systems require real robots for the assessment and planning of social interactions.
Real-time human participation in the virtual HRI scenario is our motivation.

From the viewpoint of platforms, several open source platforms for artificial intelligence (AI) agent in indoor environments, such as Malmo\,\cite{Malmo2016}, MINOS\,\cite{MINOS2017}, AI2THOR\,\cite{AI2THOR2017}, have been proposed.
The projects provide a free and open software platform that enables general users to participate in the research and development of intelligent agent systems, in which the difficulty in building robot agent and 3D environment models is removed.
Since the concept of the systems is the easy development of an autonomous agent system, the control of AI agents is limited to send simple commands by script, without using Robot Operating System (ROS).
Additionally, the systems do not support real-time interaction between real users and AI agents because the main target is the interaction between AI agents and the environments.

Another related keyword is the digital twin\,\cite{DigitalTwin:ElSaddik2018}, which creates a digital replica of the real world, including humans, robot agents, and the environment.
One of the applications of the digital twin in HRI is the investigation and optimization of the interaction/interface system and design of the robot systems.
The field of manufacturing engineering recently focuses on this technology, and several software systems are proposed\,\cite{Kuts2019,Bilberg2019}.
However, the main focus of these attempts is the real-time replication of the physical world.
The real-time participation of humans in the virtual world is not well discussed and developed.

In terms of the datasets, various multimodal behavioral datasets have been proposed toward HRI datasets.
KIT motion dataset\,\cite{Plappert2016} provides a combination of the modality of motion pattern and natural language description. Movie datasets\,\cite{Regneri2013,Sigurdsson2016} of human activity in the daily life environment have been proposed.
In terms of the conversation, collection of the utterance and evaluation of the communication in navigation task have introduced\,\cite{MacMahon2006,Vries2018,Mei2016}.
Although it is useful for the evaluation of behavior recognition and expansion to collaboration with research on natural language processing, the range of combinations of modality has a limitation.
Since it also requires dedicated experimental environments and equipments to add more data, growing the open dataset in the research community is difficult.

We direct our motivation to the realization of an open platform to collect and leverage multimodal-interaction-experience data that were collected in daily-life environments that require embodied social interaction.

    \begin{threeparttable}[t]
    \caption{Functions and limitations of related systems}
    \label{tb_diff_platforms}
    \footnotesize
    \begin{center}
      \begin{tabular}{|c|c|c|c|c|c|c|}
        \hline
        \rowcolor[gray]{0.7}
        \raisebox{0.5em}{\textbf{Platform}} &
        \raisebox{0.5em}{\textbf{Graphics}} &
        \shortstack{\\ \textbf{Physics/}\\ \textbf{dynamics}} &
        \raisebox{0.5em}{\textbf{3D model format}} &
        \shortstack{\\ \textbf{Ready-made}\\ \textbf{model/environment}} &
        \shortstack{\\ \textbf{Robotic}\\ \textbf{middleware}} &
        \shortstack{\\ \textbf{Immersion}\\ \textbf{of human}} \\
        \hline
        
        \rowcolor[gray]{0.95}
        \raisebox{0.0em}{\shortstack{\\ \textbf{Gazebo}\\ \cite{Gazebo_Koenig2004}}} &
        \raisebox{0.5em}{OGRE} &
        \raisebox{0.0em}{\shortstack{ODE, Bullet,\\ Simbody, DART}} &
        \raisebox{0.0em}{\shortstack{SDF/URDF, STL,\\ OBJ, Collada}} &
        \raisebox{0.0em}{\shortstack{40+ robot models,\\7 competitions}} &
        \raisebox{0.5em}{ROS} &
        \raisebox{0.5em}{Not supported} \\ 
        \hline
        \rowcolor[gray]{0.95}
        \raisebox{0.5em}{\shortstack{\textbf{USARSim}\\ \cite{Lewis2007}}} &
        \multicolumn{2}{c|}{\raisebox{1.0em}{Unreal Engine 2}} &
        \raisebox{1.0em}{Unknown} &
        \raisebox{0.0em}{\shortstack{\\ 5+ robot modes,\\ RoboCup Rescue,\\ RoboCup Soccer}} &
        \raisebox{0.5em}{\shortstack{Player,\\ ROS}} &
        \raisebox{1.0em}{Not supported} \\ 
        \hline
        \rowcolor[gray]{0.95}
        \raisebox{0.5em}{\shortstack{\textbf{V-REP}\\ \cite{VREP_Rohmer2013}}} &
        \raisebox{1.0em}{OpenGL} &
        \raisebox{0.5em}{\shortstack{Bullet ,ODE,\\ Vortex, Newton}} &
        \raisebox{0.0em}{\shortstack{\\ OBJ, STL, DXF,\\ 3DS, Collada,\\ URDF}} &
        \raisebox{1.0em}{\shortstack{30+ robot models}} &
        \raisebox{1.0em}{ROS} &
        \raisebox{1.0em}{Not supported} \\         
        \hline
        \rowcolor[gray]{0.95}
        \raisebox{0.5em}{\shortstack{\textbf{Choreonoid}\\ \cite{Nakaoka2012}}} &
        \raisebox{1.0em}{OpenGL} &
        \raisebox{0.0em}{\shortstack{\\ AIST engine,\\ ODE, Bullet,\\ PhysX}} &
        \raisebox{1.0em}{\shortstack{Body, VRML}} &
        \raisebox{1.0em}{\shortstack{A few robot models}} &
        \raisebox{1.0em}{OpenRTM} &
        \raisebox{1.0em}{Not supported} \\         
        \hline
        \rowcolor[gray]{0.95}
        \raisebox{0.0em}{\shortstack{\\ \textbf{Open-HRP}\\ \cite{OpenHRP_Kanehiro2004}}} &
        \raisebox{0.5em}{Java3D} &
        \raisebox{0.5em}{ODE, Bullet} &
        \raisebox{0.5em}{VRML} &
        \raisebox{0.5em}{\shortstack{A few robot models}} &
        \raisebox{0.5em}{OpenRTM} &
        \raisebox{0.5em}{Not supported} \\         
        \hline        
        \rowcolor[gray]{0.95}
        \raisebox{0.5em}{\shortstack{\textbf{Webots}\\ \cite{Webots_Michel2004}}} &
        \raisebox{0.5em}{\shortstack{WREN\\ (OpenGL)}} &
        \raisebox{1.0em}{ODE} &
        \raisebox{0.5em}{\shortstack{WBT, VRML,\\ X3D}} &
        \raisebox{0.0em}{\shortstack{\\ 50+ robot models,\\ 500+ objects,\\ 6 environments}} &
        \raisebox{0.5em}{\shortstack{ROS,\\ NaoQI}} &
        \raisebox{1.0em}{Not supported} \\         
        \hline       
        \rowcolor[gray]{0.95}
        \raisebox{0.0em}{\shortstack{\\ \textbf{OpenRAVE}\\ \cite{diankov2008_openrave}}} &
        \raisebox{0.0em}{\shortstack{Coin3D,\\ OpenSceneGraph}} &
        \raisebox{0.5em}{ODE, Bullet} &
        \raisebox{0.0em}{\shortstack{XML, VRML,\\ OBJ, Collada}} &
        \raisebox{0.5em}{10+ robot models} &
        \raisebox{0.0em}{\shortstack{ROS,\\ YARP}} &
        \raisebox{0.5em}{Not supported} \\         
        \hline

        \rowcolor[rgb]{0.95,0.95,1.0}
        \raisebox{0.0em}{\shortstack{\\ \textbf{MINOS}\\ \cite{MINOS2017}}} &
        \raisebox{0.5em}{WebGL} &
        \raisebox{0.5em}{N/A} &
        \raisebox{0.5em}{Unknown} &
        \raisebox{0.0em}{\shortstack{SUNCG,\\ Matterport3D}} &
        \raisebox{0.5em}{N/A} &
        \raisebox{0.5em}{Not supported} \\         
        \hline
        \rowcolor[rgb]{0.95,0.95,1.0}
        \raisebox{0.0em}{\shortstack{\\ \textbf{Project Malmo}\\ \cite{Malmo2016}}} &
        \multicolumn{2}{c|}{\raisebox{0.5em}{Minecraft}} &
        \raisebox{0.5em}{Unknown} &
        \raisebox{0.5em}{\shortstack{MARL\"{O} competition}} &
        \raisebox{0.5em}{N/A} &
        \raisebox{0.5em}{Not supported} \\
        \hline        
        \rowcolor[rgb]{0.95,0.95,1.0}
        \raisebox{0.5em}{\shortstack{\textbf{AI2THOR}\\ \cite{AI2THOR2017}}} &
        \multicolumn{2}{c|}{\raisebox{1.0em}{Unity}} &
        \raisebox{0.0em}{\shortstack{\\ FBX, Collada,\\ 3DS, DXF,\\ OBJ, …}} &
        \raisebox{0.5em}{\shortstack{200+ environments,\\ 2600+ objects}} &
        \raisebox{1.0em}{N/A} &
        \raisebox{1.0em}{Not supported} \\         
        \hline
        \rowcolor[rgb]{0.95,0.95,1.0}
        \raisebox{0.5em}{\shortstack{\textbf{VirtualHome}\\ \cite{Puig2018_virtualhome}}} &
        \multicolumn{2}{c|}{\raisebox{1.0em}{Unity}} &
        \raisebox{0.0em}{\shortstack{\\ FBX, Collada,\\ 3DS, DXF,\\ OBJ, …}} &
        \raisebox{0.0em}{\shortstack{6 environments,\\ 350+ object models,\\ Knowledge base}} &
        \raisebox{1.0em}{N/A} &
        \raisebox{1.0em}{Not supported} \\
        \hline
        
        \rowcolor[rgb]{1.0,1.0,0.95}
        \raisebox{0.0em}{\shortstack{\\ \textbf{DeepMind Lab}\\ \cite{Beattie2016_DeepMindLab}}} &
        \multicolumn{2}{c|}{\raisebox{0.5em}{Quake III Arena}} &
        \raisebox{0.5em}{Unknown} &
        \raisebox{0.5em}{\shortstack{Several games}} &
        \raisebox{0.5em}{N/A} &
        \raisebox{0.5em}{Not supported} \\         
        \hline
        \rowcolor[rgb]{1.0,1.0,0.95}
        \raisebox{0.0em}{\shortstack{\\ \textbf{OpenAI Gym}\\ \cite{Brockman2016_OpenAI_Gym}}} &
        \multicolumn{2}{c|}{\raisebox{0.0em}{\shortstack{MuJoCo, Atari, Box2D,\\ 15+ simulation environments}}} &
        \raisebox{0.5em}{Unknown\tnote{a}} &
        \raisebox{0.5em}{Unknown\tnote{a}} &
        \raisebox{0.5em}{N/A\tnote{b}} &
        \raisebox{0.5em}{Not supported} \\
        \hline        

        \rowcolor[rgb]{0.95,1.0,0.95}
        \raisebox{0.0em}{\shortstack{\\ \textbf{iCub-HRI}\\ \cite{iCub-HRI_Fischer2018}}} &
        \multicolumn{2}{c|}{\raisebox{0.5em}{N/A}} &
        \raisebox{0.5em}{N/A} &
        \raisebox{0.5em}{1 robot model} &
        \raisebox{0.5em}{YARP} &
        \raisebox{0.5em}{Not supported} \\
        \hline
        
        \rowcolor[rgb]{1.0,0.95,0.95}
        \raisebox{0.0em}{\shortstack{\\ \textbf{SIGVerse (Ver.2)}\\ \cite{Inamura2010}}} &
        \raisebox{0.5em}{OGRE} &
        \raisebox{0.5em}{ODE} &
        \raisebox{0.5em}{X3D, VRML} &
        \raisebox{0.5em}{A few robot model} &
        \raisebox{0.5em}{N/A} &
        \raisebox{0.5em}{Supported} \\
        \hline
        \rowcolor[rgb]{1.0,0.95,0.95}
        \raisebox{1.0em}{\shortstack{\textbf{SIGVerse (Ver.3)}}} &
        \multicolumn{2}{c|}{\raisebox{1.0em}{Unity}} &
        \raisebox{0.0em}{\shortstack{\\ FBX, Collada,\\ 3DS, DXF,\\ OBJ, …}} &
        \raisebox{0.0em}{\shortstack{5+ robot models,\\ 200+ objects,\\ 40+ environments}} &
        \raisebox{1.0em}{ROS} &
        \raisebox{1.0em}{Supported} \\
        \hline
      \end{tabular}
      \begin{tablenotes}
        \item[a] It depends on the used simulation environment.
        \item[b] ROS is supported only in a third-party environment gym-gazebo.
      \end{tablenotes}
    \end{center}
    \end{threeparttable}

\section{SIGVerse: a cloud-based VR platform}

In this section, we introduce the concept of a could-based VR system to accelerate the data collection in HRI applications.

For example, collaborative cooking tasks and dialogue-management systems dealing that address vague utterances and gestures, and gazing behaviors are assumed to be target situations that involve HRI scenarios.
In these situations, a robot must observe and learn social behaviors of the humans with which it interacts and solve ambiguities that are based on past interaction experience.
In these complex environments, the robot should collect the following multimodal data:

\begin{enumerate}
    \item Physical motion/gestures during interaction (including gaze information)
	\item Visual information (i.e., image viewed by the agents)
	\item Spatial information (i.e., positions of agents and objects)
	\item Voice interaction (i.e., utterance of agents)    
\end{enumerate}

Additionally, the following functions must be provided:
\begin{itemize}
	\item[(i)] Users are able to login to avatars in the VR environment from anywhere
    \item[(ii)] Multiple-users are able to simultaneously login to the same VR scene via the Internet
    \item[(iii)] Recording and replaying time-series multimodal interaction data
	\item[(iv)] Attaching control programs of real robots to virtual robots    
\end{itemize}

The functions (i) to (iii) are based on the real-time participation of humans in the virtual environment, which was not discussed in conventional robot simulators.
The function (iv) is required for the efficient development of robot software, which could be used in both real and virtual environments.
Thus, the support of robotic middleware is essential. The function (iii) requires good quality of graphics function and computational power for the physic simulation.
The reusability of the ready-made 3D model of the robot and daily life environment is also essential to construct a variety of virtual environments for HRI experiments.
Since we have several types of data format for the robot model, compatibility is another important function for efficient development.
Table.\ref{tb_diff_platforms} shows the performance of existing related systems from the viewpoint of these required functions.

Our previous system (SIGVerse ver.2)\,\cite{Inamura2010} has been utilized for studies such as analysis of human behavior\,\cite{Karinne2014}, learning of spatial concepts\,\cite{Taniguchi2016}, and VR-based rehabilitation\,\cite{Inamura:AR2017}.
These studies employed multimodal data (1) to (4) and functions (i) to (iii); however, the re-usability of conventional SIGVerse is restricted due to its application programming interface (API).
Therefore, the system needs to keep functions (i) to (iii) and realize function (iv).
The next subsections present the software architecture that realizes these functions.

\subsection{Architecture of the SIGVerse}

\begin{figure}[th]
  \begin{center}
  \includegraphics[width=15cm]{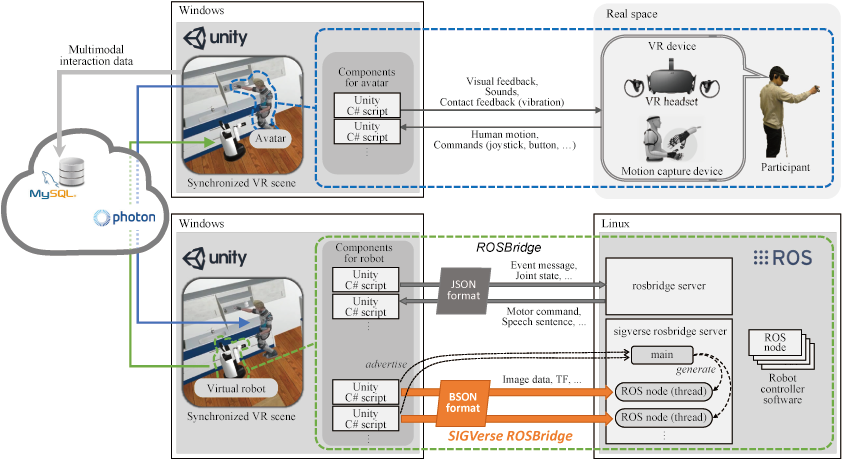}
  \end{center}
  \caption{Software architecture of the SIGVerse system}\label{fig:architecture}
\end{figure}

The detailed architecture of SIGVerse (ver.3), including a participant and a robot, is shown in Fig.\,\ref{fig:architecture}.
SIGVerse is a server/client system that is based on a third-party networking technology (i.e., Photon Realtime).
The server and clients have the same scene, which is composed of 3D object models such as avatars, robots, and furniture.
By communicating information of registered objects via the Internet, the events in each scene can be synchronized.

The participant can login to an avatar via VR devices such as head-mounted displays (HMDs), hand-tracking controllers, audio devices, and motion capture devices.
According to the input from these VR devices, behavior of the participant is reflected on the avatar by Unity scripts.
Perceptual information such as perspective visual feedback is provided to the participant.
Thus, the participant can interact with the virtual environment in a manner that is similar to a real environment.

The proposed VR simulation system has a bridging mechanism between ROS and Unity.
Software for virtual robot control can be reused in real robots without modification, and vice versa.

The information for reproducing multimodal interaction experiences is stored on a cloud server as a large-scale dataset of embodied and social information.
By sharing this information, users can reproduce and analyze the multimodal interaction after the experiment.

\subsection{Mechanism for connecting ROS and Unity}
To control a robot in a VR environment, sensory feedback and robot commands should be passed between Unity scripts and ROS nodes.
The most important factor in realizing the integration of ROS and Unity is the communication protocol between them.
Software systems for bridging ROS and Unity have been proposed by Hu et al.\,\cite{ROSUnitySim_Hu2016} and Downey et al.\,\cite{Downey2014}.
For these systems, motor commands and sensor information are transferred using rosbridge.
However, if users attempt to send a massive amount of sensor information, such as camera images, from Unity, a bottleneck on the transfer speed is created.
Previous works\,\cite{ROSUnitySim_Hu2016,Downey2014} did not discuss how to transfer camera images in realtime; accordingly, a new technique for realizing real-time transfer based on the binary JavaScript object notation (BSON) format using a TCP/IP connection is proposed in the following section.

As a ROS functionality, a rosbridge framework provides a JavaScript object notation (JSON) API and a WebSocket server for communicating between a ROS program and an external non-ROS program.
JSON is a text-based data exchange format that represents pairs of keywords and values.
Although the rosbridge protocol covers sending and receiving ROS messages, its performance for parsing large JSON data, such as images, is insufficient for satisfying real-time sensor feedback.
For this reason, a specific server (sigverse\_rosbridge\_server) for communicating large data volumes was implemented.
For speeding up communication, the BSON format was employed instead of JSON.
BSON is a binary-encoded serialization with a JSON-like format.
The use of BSON offers the following advantages: the communication data size is reduced to less than that of text-based data;
a conversion process between text and binary is not required;
and data are represented as key-value pairs that are compatible with ROS messages.
When ROS messages are advertised by Unity scripts, the main thread of the sigverse\_rosbridge\_server generates a new thread for each topic as a ROS node.
Each thread receives ROS messages from the Unity scripts and publishes them in the ROS nodes of the robot controller as ROS topic messages.

Siemens and the ROS community has proposed the software module ROS\# to support the communication between ROS and Unity\footnote{https://github.com/siemens/ros-sharp}.
Since the ROS\# employs the WebSocket to send and receive a control command and status of the robot, realizing the real-time data transfer is difficult by the ROS\#.

\begin{figure}[th]
  \begin{center}
  \includegraphics[width=10cm]{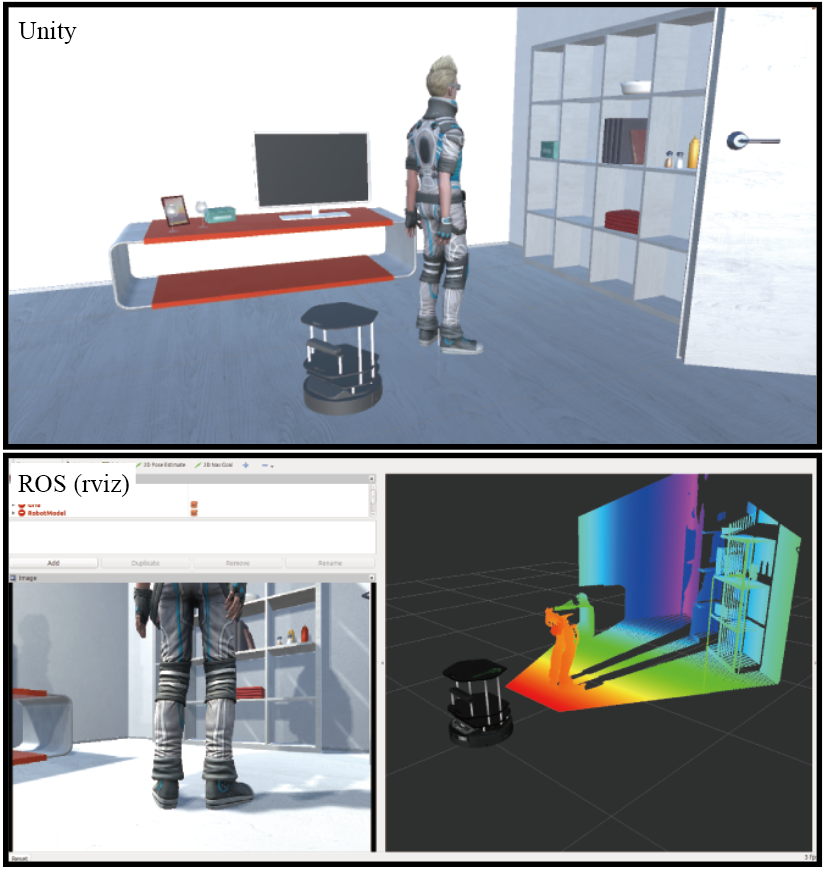}
  \end{center}
  \caption{Virtual RGB image and depth data communicated from Unity to ROS. The top figure shows a VR scene, including a robot with an RGB-D sensor, and the bottom figure shows an RGB image and the depth data received in ROS.}
  \label{fig:sensor_data}
\end{figure}

We evaluated the data transfer performance\,\cite{Mizuchi2017} to compare the proposed method and conventional JSON based method.
An experimental condition shown in Fig.\ref{fig:sensor_data} was used for the investigation where a mobile robot tracks walking person using an RGB-D sensor.
The PC with Xeon E5-2687W CPU and GeForce GTX TITAN X GPU was used in this evaluation.
The size of a raw RGB-D frame was 1.5 MB.
The averages of frequency of RGB-D data are listed in Table\,\ref{tb_frequency}.
The JSON communication was insufficient for satisfying the real-time requirement for HRI, even if a high-end computer was employed.

\renewcommand{\arraystretch}{1.2}
\begin{table}[tb]
\caption{Frequencies of virtual RGB-D data depending on protocols}
\label{tb_frequency}
\begin{center}
  \begin{tabular}{|c|c|}
    \hline
    \textbf{WebSocket with JSON} & \textbf{TCP/IP with BSON} \\
    (Conventional rosbridge) & (sigverse\_rosbridge) \\
    \hline
    0.55 [fps] & 57.60 [fps] \\
    \hline
  \end{tabular}
\end{center}
\end{table}
\renewcommand{\arraystretch}{1.0}

\subsection{Configuration of the cloud-based VR space for HRI}

To enable general users to participate in an HRI experiment from an arbitrary location, we developed a cloud-based VR function on SIGVerse.
Each user logs in to an avatar by a client computer, which has a Unity process for the VR devices.
The computer for a user connects to another computer, which has another Unity process to control the behavior of the target robot.
The internal state of all VR scenes are synchronized via the Internet based on Photon Realtime, which is a universal software module to integrate different game engines. For the simple use case, each computer (Unity process) connects to a cloud server provided by Photon inc.

Fig.\,\ref{fig:architecture} also depicts the cloud configuration employed in SIGVerse.
One computer (Unity process) is assigned to each user/robot to realize the complex interaction between multi-robots and multi-users.
A mixed environment consists of virtual and real environments that are based on Augmented Reality (AR) devices, such as HoloLens and real motion capturing systems.
This usage is one of the future possibilities of the SIGVerse platform.

\begin{figure}[th]
  \begin{center}
  \includegraphics[width=10cm]{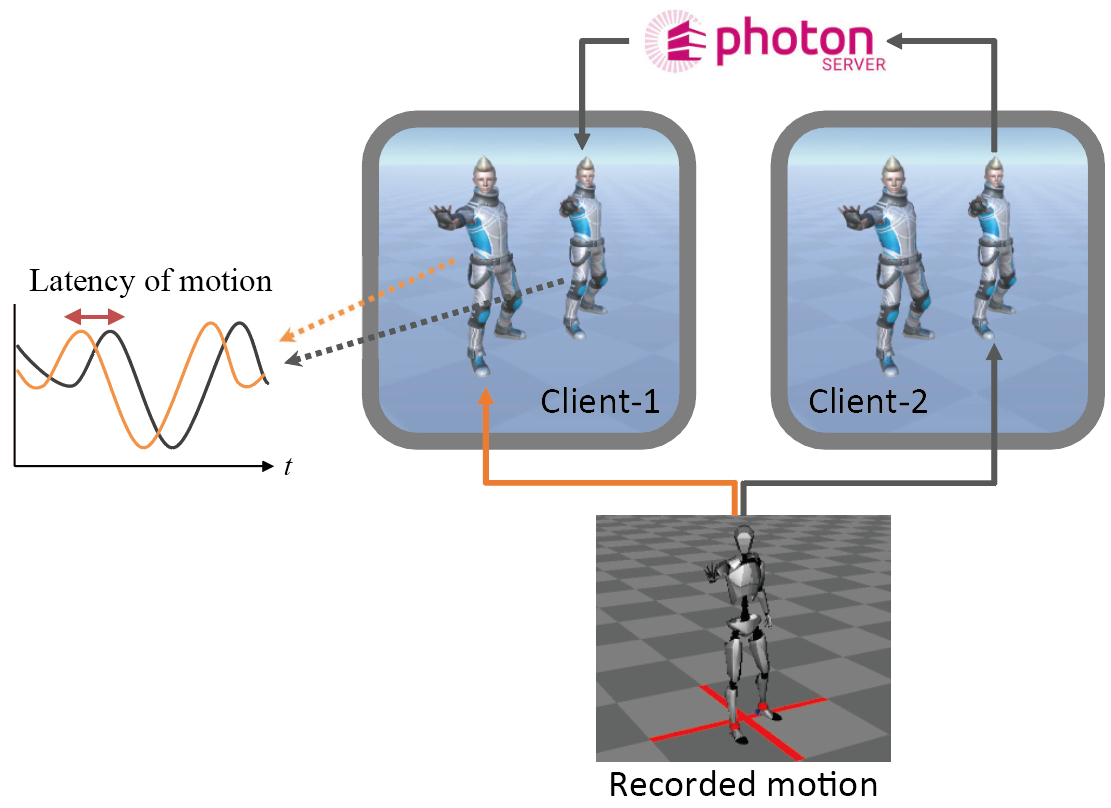}
  \end{center}
  \caption{Configuration of the latency evaluation.}
  \label{fig:latency_evaluation}
\end{figure}

We measured the latency between a local avatar and other avatars, whose motion are synchronized using a local Photon server.
The configuration of the latency evaluation is shown in Fig.\ref{fig:latency_evaluation}.
A motion player software (Axis Neuron) broadcasted pre-recorded motion data via TCP/IP connection.
As all the clients were located in the same place, the motion difference among local avatars is negligible.
The poses of the non-local avatars were synchronized with the poses of the corresponding avatars in each client.
The position and rotation of 56 joints, including the fingers, were synchronized for each avatar.
The latency was evaluated in 2-client, 4-client, and 8-client cases. 

\begin{figure}[th]
  \begin{center}
  \includegraphics[width=15cm]{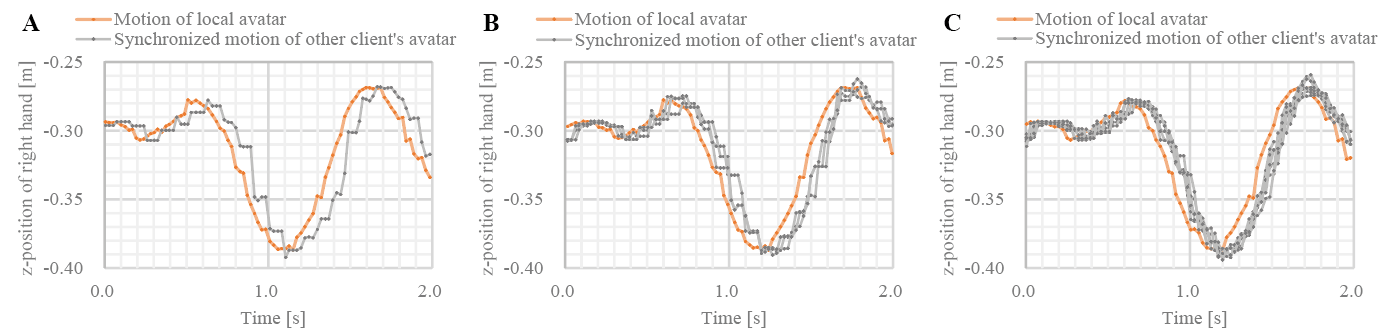}
  \end{center}
  \caption{Motions of local avatar and synchronized avatars, \textbf{(A)} is motions in 2-clients case, \textbf{(B)} is motions in 4-clients case, and \textbf{(C)} is motions of 8-clients case.}
  \label{fig:latency}
\end{figure}

The $z$-position of the right hand in each case is plotted in Fig.\,\ref{fig:latency}.
Although the motions of non-local avatars were slightly disturbed and delayed, their motions were synchronized.
The latency of synchronized motions was approximately 70~ms regardless of the number of clients.
This latency is sufficiently low to allow multimodal interaction and cooperative tasks among multiple robots and avatars in the cloud-based VR space.

\subsection{Database subsystem for the record of Human-Robot Interaction}

The SIGVerse system has a database subsystem to record multimodal HRI behavior.
The targets of the recording consist of physical and cognitive interaction.
As the physical interaction, the system records action of the robot and avatar (time-series data of joint angles and positions of the robot and avatar), and time-series data of the status of objects that are manipulated by the robot and the avatar.
As the cognitive interaction, the system records verbal conversation between the robot and avatar.
Since the recorded data could be stored in the MySQL server, a client can reproduce the physical and cognitive interaction in the VR environment from anywhere via the Internet.

Although measured sensor signals by the robot is useful information as the HRI record, the database subsystem does not record the sensor signals because the computational load for simulating and recording the sensors high.
SIGVerse provides a function for reproducing the sensor signals from the record of the behavior of the robot, avatar, and objects, instead of the direct recording of the sensor signals.

\section{Case studies}

\subsection{Robot competition}

One of the better and effective methods for evaluating the performance of HRI is a robot competition, such as RoboCup@Home\,\cite{Iocchi2015}.
Since the organization of a robot competition requires a considerable amount of time and human resources, simulation technologies were often employed to reduce the cost.
Even though RoboCup Soccer and RoboCup Rescue have simulation leagues, only RoboCup@Home does not have the simulation league.
One of the reasons and difficulties for realizing the RoboCup@Home simulation is the necessity of interaction between real users and virtual robots.
This problem could be solved by SIGVerse.

Since changing the rulebook and the competition design of RoboCup is difficult,
we organized a VR-based HRI competition in the World Robot Summit, Service Category, Partner Robot Challenge\footnote{World Robot Summit, Service Category, Partner Robot Challenge \url{http://worldrobotsummit.org/en/wrc2018/service/}} that was held in October 2018 in Tokyo\,\cite{Okada2019}.
The following subsections present two representing tasks in the competition based on the proposed system.

\subsubsection{Task 1: Interactive Cleanup}
Gesture recognition is a basic function of understanding human behavior.
Since the image data set\,\cite{Wan2016} and competition\,\cite{Escalera2013} has an important role in this area from the past several decades.
The gesture recognition functions that are required for intelligent robots that work with people include not only label recognition but also following a moving human and observation of the gesture from an easy-to-view angle.
Furthermore, recognizing the pointing target object from the spatial relationship between the object that exists in the environment and the human.

\begin{figure}[th]
  \begin{center}
  \includegraphics[width=15cm]{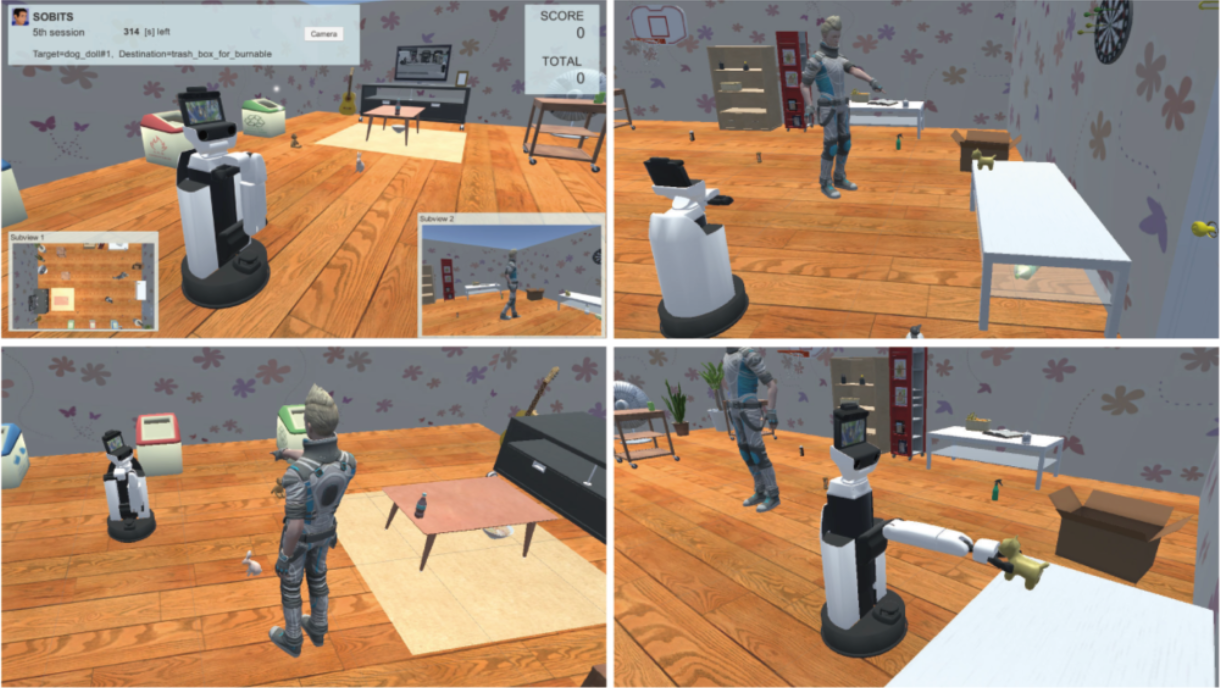}
  \end{center}
  \caption{Screenshot of the Interactive CleanUp task}
  \label{fig:interactive_cleanup}
\end{figure}

\subsubsection{Task 2: Human Navigation}
\label{sec:human_navigation}

We focus on a task named 'Human Navigation'\cite{Inamura:RoboCup2017} in which the robot has to generate friendly and simple natural language expressions to guide people to achieve several tasks in a daily environment, for evaluation of HRI in the VR environment. 
The roles of the robot and the user are opposite to the roles of those in the conventional task, such as understanding and achieving a request given by users.
The robot has to provide natural language instructions to tell a person to carry a certain target object to a certain destination, such as, “Please take the cup on the table in front of you to the second drawer in the kitchen.”
A person (test subject) logs in to an avatar in a VR space to follow the instructions.
The test subject attempts to pick up the target object and take it to the destination using virtual reality devices following the instructions from the robot.
The required time to complete the object transportation is measured and applied to calculate points.
The team that generates the easiest and most natural instructions for a person to follow will receive more points. 

Although the competition participants do not need to learn about the VR system, they should concentrate on the software development on the ROS framework.
The outcome of the developed robot software could be easily applied to a real robot system.

According to the rulebook\footnote{\url{http://worldrobotsummit.org/download/rulebook-en/rulebook-simulation_league_partner_robot_challnege.pdf}} of this competition, we evaluated the reaction of the test subjects regarding how to address the utterance of the robot.
Basic analysis of the effectiveness of the interaction is performed by the required time.
If the instruction of the robot is not friendly, the test subjects tend to be confused and exhaust a substatial amount of time to complete the task.
\begin{figure}[htbp]
  \centering
  \includegraphics[width=0.9\columnwidth,height=\textheight,keepaspectratio]{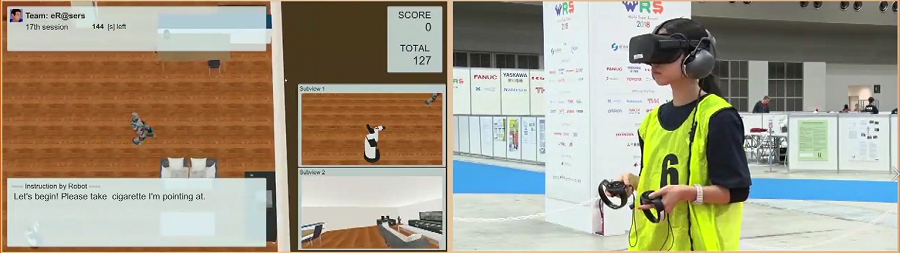}
  \includegraphics[width=0.9\columnwidth,height=\textheight,keepaspectratio]{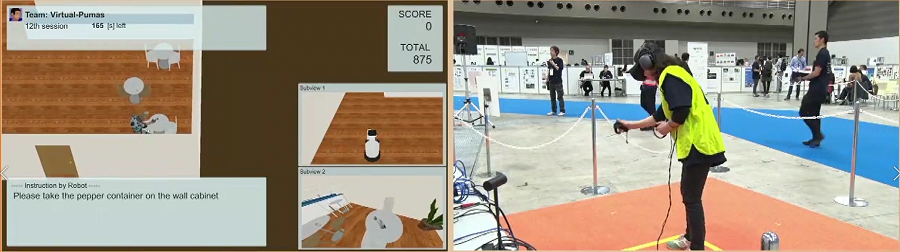}
  \caption{Screenshot of the Human Navigation task. Written informed consent was obtained from the test subject for the publication of any potentially identifiable images or data included in this article.}
  \label{fig:screenshot1}
\end{figure}

\begin{figure}[th]
  \begin{center}
  \includegraphics[width=15cm]{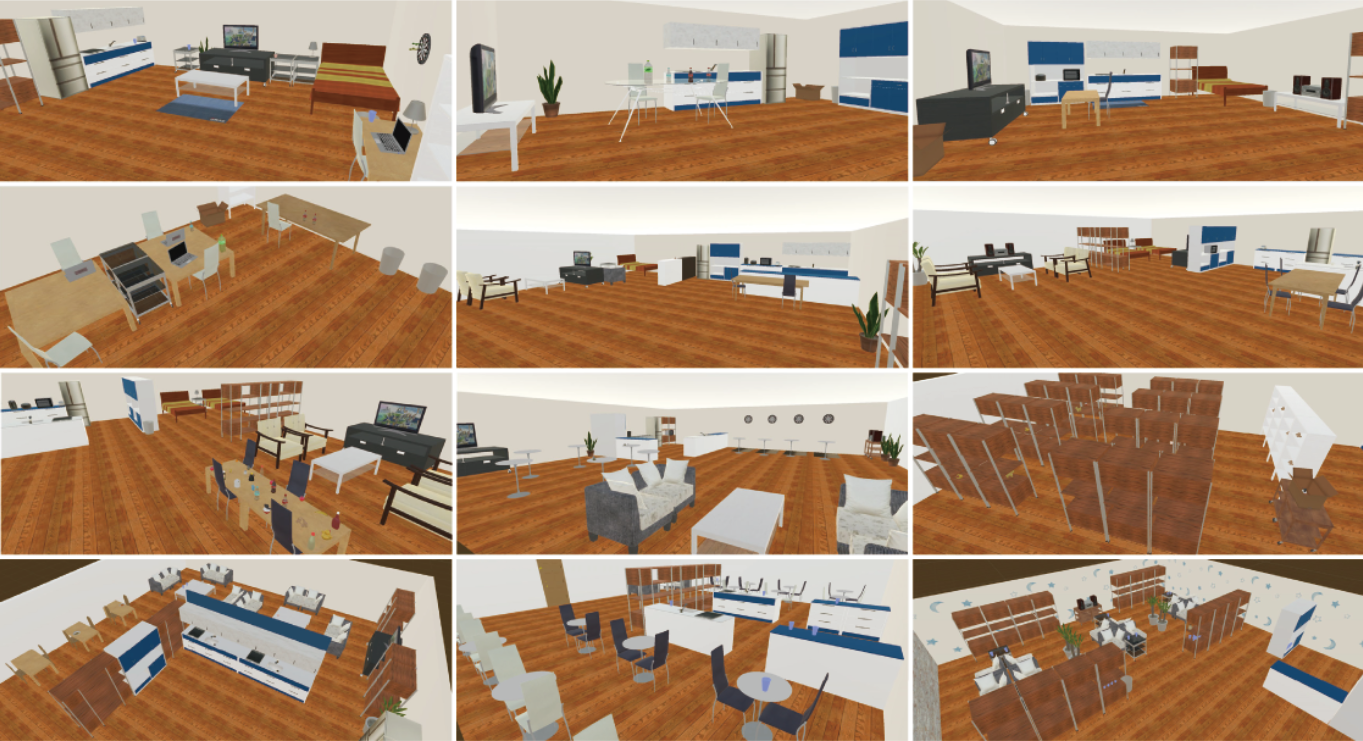}
  \end{center}
  \caption{Room layouts}
  \label{fig:layouts}
\end{figure}

\subsection{Modeling of subjective evaluation of HRI ability}

The interaction ability of mobile robots in the Human Navigation task is mainly evaluated by the required time to achieve the task.
However, other factors would be the target of evaluation such as the number of used words, frequency of the pointing gesture,
and length of the trajectory of the avatar's behavior.
Designing objective and fair criteria for the evaluation is difficult because a substantial number of factors are evaluated subjectively in a daily life environment.
Although the best evaluation method is to ask several referees to score the performance in several trials,
the human navigation was evaluated by a certain regulation, such as a positive point for 'desirable behavior' and a negative point for 'unfriendly behavior', which is described in the rulebook from a subjective viewpoint.
We have cleared this problem to extract a dominant factor for the evaluation of the interaction behavior in the HRI dataset\,\cite{Mizuchi2020}.

Not only motion and utterance by the robot but also social and embodied behavior of the avatar is recorded in the dataset, such as confused behavior when the avatar receives strange guidance and meaningless behavior by misunderstanding the guidance.
Even if the evaluation method in the competition failed to focus on a certain important factor, the evaluation system can analyze the
missing parameter by reproducing the interaction behavior in the VR environment, for example, an image sensor observed by the robot in a past competition.

After recording the human-robot interaction behavior, we asked the third person, who was not concerned with the competition, to evaluate the quality of the interaction.
Since the HRI behavior was played back in the VR  system, the test subjects could check the detailed behavior by changing the viewpoint, rewinding the past behavior, and repeating the important scene, similar to the recent technology named Video Assistant Referee, which is used in football games.
A 5-point Likert scale questionnaire is utilized for the evaluation.
We collected evaluation data in 196 sessions using 16 test subjects.
To estimate the evaluation results as the objective variable, we selected 10 parameters as the explanatory variable, such as the required time to complete the task, number of utterances from the robot and accumulated angle of the change in the gaze direction.
We conducted multiple regression analysis to estimate the evaluation by the third person's assessment.

The results showed that the subjective evaluation result can be estimated by the objective parameter, which is related to the physical and social interaction between humans and robots.
We compared the quality of the evaluation criterion between the result of regression analysis and the rulebook, which was used in the robot competition.
The ranking determined by the evaluation criterion, which was calculated by our method, differed from the ranking determined by the rulebook and resembled the ranking determined by a third party.
Since the parameters are easily measured in the VR environment, an automatic evaluation of the HRI behavior in the robot competition becomes easier without a subjective assessment by the referee.
Since recording the behavior in the real robot competition is too difficult, the proposed VR system would be one of the key technologies for evaluating the performance of the social robot in the real environment.

\subsection{Motion learning by demonstration}
Another advantage of the proposed system is the ease of behavior collection for robot learning.
Motion learning by demonstration is one of the major applications in robot learning that requires human motion and an interaction process between a human and the environment.

Bates et al. have proposed a virtual experimental environment for learning the dish washing behavior\,\cite{Bates2017} based on the SIGVerse system.
The test subjects logged into the VR environment and operated dishes and a sponge by hand devices, such as Oculus Touch.
The motion patterns and state transition history of the related objects are recorded; and a semantic representation was learnt from the collected data.
Finally, a robot imitates the dish washing behavior in the real world with reference to the semantic representation even if the robot never observes human behavior in the real world.

Vasco et al also proposed a motion learning method based on our system\,\cite{Vasco2019}.
They aimed to make the robot system learn the motion concept, which includes not only motion patterns but also related information, such as operated tools and objects by the motion and location information where the motion is performed.
Our system was used to collect the motion data in a variety of situations in a short time.

In both applications, the target of the learning is not only the motion patterns but the interaction process between the environments and objects, and the effect on the environment by the performance is the focus.
The virtual environment provides an easy recording function of motion patterns and the state transition process of virtual environments and objects.
Although the motion capture system and object tracking functions are available in the real world, the cost to build the real field environment is still expensive for researchers.

Since our motivation is an extension of the opportunity to collect the target behavior, the system has an observation function for behavior, which is not performed by real users.
Behaviors performed by a virtual robot was transferred to another robot via SIGVerse in \cite{Karinne2014}.
This flexibility is another advantage of the SIGVerse system.

\subsection{VR based neurorehabilitation}

Recent VR technology is employed in several rehabilitation therapies, such as mirror therapy and imitation therapy.
Mirror therapy is one of the famous therapies for phantom limb pain that arises in patients who lost a limb in an accident.
Conventional mirror therapy employs a physical mirror to display the intact limb for the patients, whereas VR rehabilitation systems display the intact limb as a virtual avatar instead of using a mirror.

\begin{figure}[th]
  \begin{center}
  \includegraphics[width=10cm]{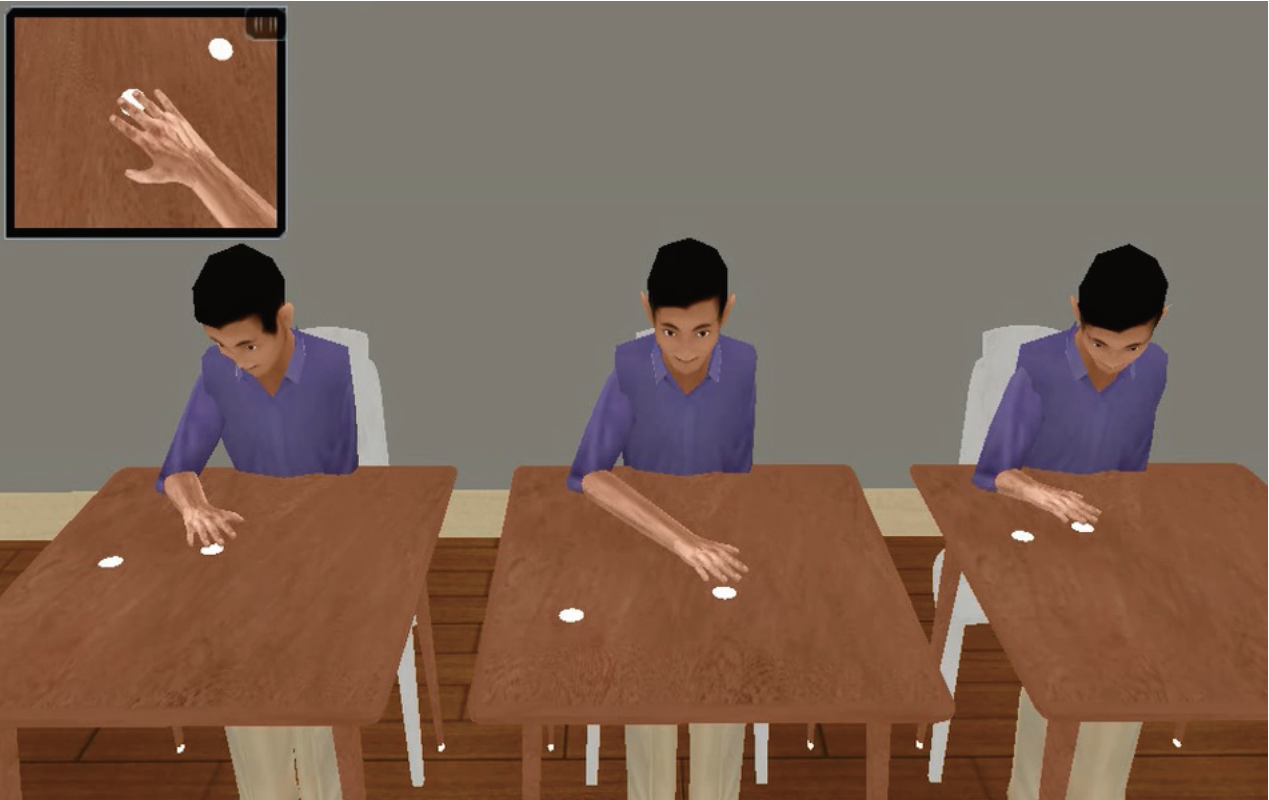}
  \end{center}
  \caption{Patient avatar in the neurorehabilitation for telescoping phantom limb.}
  \label{fig:three_length_limbs}
\end{figure}

The limitation of a conventional VR system for mirror therapy is that addressing the phenomenon referred to as “telescoping” is difficult.
Patients often feel an usual length or shape of a limb, such as short/long limb, bended limb, limb and sunk into body.
In this case, mirror therapy does not work well because the appearance of a limb on the VR avatar differs from the phantom limb representation in the brain.
Thus, we design the customized limb based on the subjective feeling of the telescoped limb shown in Fig.\ref{fig:three_length_limbs}.

One of the difficulties in neurorehabilitation is the evaluation of the therapy method with the collecting of a large amount of reaction data from patients.
Our system has the potential to be applied to the clinical field in which patients are unable to go to the hospital frequently.
Even if patients stay in their homes, our system enables therapists to deliver an appropriate rehabilitation program via the Internet and VR devices and collect the reaction behaviors of patients via the cloud system\,\cite{Inamura:AR2017}.
Originally, our system is designed to collect datasets in HRI experiments; however, almost the same framework is employed in therapist-patient interaction.
From the viewpoint of a cloud-based system, this framework has the potential to realize a home rehabilitation system.

\section{Discussion}
\subsection{Limitation}
Since a physics simulation is performed in the Unity system, a complex simulation is one of the limitations, such as friction force, manipulation of soft materials, and fluid dynamics.
Additionally, a standard 3D shape model of the robot, such as the URDF utilized in Gazebo, is not easily imported to the SIGVerse system due to the format of the mechanism description.
Currently, we need to modify the URDF format for use in SIGVerse by manual work.

Design of software modules to control the virtual robot is another limitation.
The controller modules in robot simulators, such as Gazebo, are often provided by the manufacturer of the robot, which is executed as a process on Ubuntu.
However, we have to port the robot controller into C\#, which should be executed on Unity.
The cost for the porting should be discussed when general users employ SIGVerse system.
Four types of robot: HSR\,\cite{Yamamoto2018}, Turtlebot2, Turtlebot3, PR2\,\cite{PR2:Wyrobek2008} and TIAGo\,\cite{TIAGo:Pages2016} are currently provided by the developer team.

The advantage of SIGVerse is that the test subject can easily participate in experiments in the VR environment.
However, an autonomous agent module that acts in the VR environment without real test subject/users is not realized.
A future research direction is to construct an autonomous agent module that is based on the analysis of a HRI dataset.
The original dataset captured in the HRI experiments and augmented datasets, which could be generated in the VR environment, should be employed in the construction process based on machine learning techniques\,\cite{Goutsu2019}.

The gap between the real world and the virtual environment often becomes a target of discussion.
The robot motion that is controlled in the virtual simulator is one of the frequent arguments in robotics research.
Furthermore, the cognitive behavior of test subjects is another argument.
We have investigated the difference in human behavior that is derived from the condition of the field of view (FOV) of HMD\,\cite{Mizuchi2018}.
Distance perception in the VR environment\,\cite{Phillips2010} will be another problem for evaluating the HRI in the VR.
Thus the appropriate design of the VR environment should be discussed, in which test subjects can behave in the same way as they will in the real world.

\subsection{Future direction of the VR platform}
The current SIGVerse platform operates only in a VR environment; however, by using the AR function, virtual robot agents could expand to the situation where they interact with humans in the real world.
The Human Navigation task described in Section\,\ref{sec:human_navigation} is an example where the AR system can apply for the improvement in the intelligence of service robots in the real world.
This function will be addressed as a future task since an extensive range of applications can be expected by adjusting the boundary between the virtual environment and the real-world environment according to the situation and tasks.

In the field of image processing such as MNIST and ImageNet, many datasets exist for object recognition using machine learning, and a platform that can fairly and objectively evaluate the performance of the algorithm proposed by each researcher is provided.
We can use several datasets related to human activities.
Video clips\,\cite{Patron-Perez2012} and motion capture data\,\cite{Xia2017} for human movements, natural language sentences to describe the movements\,\cite{Plappert2016}, and conversational data to guide the user to the destination\,\cite{Vries2018} can be employed.
However, no dataset in the HRI field contains conversation to manipulate the object and navigate in a complicated daily life environment.
These datasets are indispensable for promoting the research of interactive intelligent robots in the future, and the VR platform described in this paper has the position as the foundation.

\section{Conclusion}
We developed a software platform to accelerate HRI research based on an integration of the ROS and Unity framework.
Due to the exclusive ROS bridging mechanism, which is based on BSON communication, real-time interaction among virtual robots controlled by ROS and real humans who use an application on Unity is realized.
We also introduced and showed the feasibility of the platform in four applications: robot competition, evaluation of HRI, motion learning by demonstration and neurorehabilitation.
In these applications, we confirmed that the data collection did not take substantial amount of time, and effective experiments were available in the VR environment.

As we mentioned in the introduction, future intelligent robots require the ability to deepen social behavior in a complex real world.
For this purpose, the dataset applied for social behavior learning and performance evaluation should be established.
A simulation environment that allows autonomous robots and real humans to interact with each other in real-time is important for both the preparation of these datasets and the establishment of an objective evaluation of HRI.
The proposed system, which combines a VR system and a robot simulation, becomes an important approach for realizing this HRI dataset.

\section*{Acknowledgments}
The authors would like to thank Hiroki Yamada to support the development of the cloud-based VR platform as a software technician.

\bibliographystyle{unsrt}  


\end{document}